# Multimodal Cardiovascular Risk Profiling Using Self-Supervised Learning of Polysomnography

Zhengxiao He[1], Huayu Li[1], Geng Yuan[2], William D.S. Killgore[3,4], Stuart F. Quan[5,6], Chen X. Chen[4,7], Ao Li[1,4]

[1] Department of Electrical and Computer Engineering, University of Arizona, Tucson, AZ, USA

[2] School of Computing, University of Georgia, Athens, GA, USA

[3] Department of Psychiatry, University of Arizona, Tucson, AZ, USA

[4] BIO5 Institute, University of Arizona, Tucson, AZ, USA

[5] Department of Medicine, University of Arizona, Tucson, AZ, USA

[6] Harvard Medical School and Brigham and Women's Hospital, Boston, MA, USA

[7] College of Nursing, University of Arizona, Tucson, AZ, USA

*Corresponding Author: Ao Li, PhD

Department of Electrical and Computer Engineering, University of Arizona, Tucson, AZ, USA

Email: aoli1@arizona.edu

**Abstract**

**Study Objectives**: Polysomnography (PSG) provides a comprehensive assessment of brain, cardiac, and respiratory activity during sleep. While it is widely used for diagnosing sleep disorders, its potential to assess future health risks has not been fully explored. This study aimed to develop and evaluate an interpretable framework to identify physiological patterns in PSG data linked to cardiovascular disease (CVD) outcomes, without relying on manual annotations (e.g., sleep stages).

**Methods**: We developed a self-supervised deep learning model that extracts meaningful patterns from multi-modal signals (Electroencephalography (EEG), Electrocardiography (ECG), and respiratory signals). The model was trained on data from 4,398 participants. Projection scores were derived by contrasting embeddings from individuals with and without CVD outcomes. External validation was conducted in an independent cohort with 1,093 participants. The source code is available on https://github.com/miraclehetech/sleep-ssl.

**Results**: The projection scores revealed distinct and clinically meaningful patterns across modalities. ECG-derived features were predictive of both prevalent and incident cardiac conditions, particularly CVD mortality. EEG-derived features were predictive of incident hypertension and CVD mortality. Respiratory signals added complementary predictive value. Combining these projection scores with the Framingham Risk Score consistently improved predictive performance, achieving area under the curve values ranging from 0.607 to 0.965 across different outcomes. Findings were robustly replicated and validated in the external testing cohort.

**Conclusion**: Our findings demonstrate that the proposed framework can generate individualized CVD risk scores directly from PSG data. The resulting projection scores have the potential to be integrated into clinical practice, enhancing risk assessment and supporting personalized care.

**Keywords:** self-supervised learning; cardiovascular disease; polysomnography; latent representation; risk assessment

**Statement of Significance**

This is one of the first studies to apply a self-supervised framework for cardiovascular risk profiling from PSG data. By transforming EEG, ECG, and respiratory signals into interpretable projection scores, we identified physiological markers predictive of multiple cardiovascular outcomes. These risk profiles, combined with traditional risk scores, significantly improved predictive performance across both internal and external cohorts. Our findings highlight the untapped potential of PSG signals beyond sleep staging, providing an interpretable, scalable, and clinically actionable approach for personalized cardiovascular risk stratification.

# 1 Introduction

The critical role of sleep in cardiovascular health has become increasingly evident in recent decades. Numerous epidemiological studies have consistently linked poor sleep quality or insufficient sleep duration to elevated risks of cardiovascular conditions, including hypertension, coronary artery disease, and stroke[1–3]. Despite these associations, conventional approaches for quantifying sleep's impact on cardiovascular risk often rely on summary metrics from polysomnography (PSG), which can obscure the rich physiological dynamics present in raw PSG signals.

PSG, the gold standard for sleep assessment, enables high-resolution, synchronized recordings of brain, cardiac, and respiratory activity throughout the night[4]. Traditional summary PSG-derived features, such as sleep efficiency, REM proportion, arousal index, and apnea-hypopnea index (AHI), have been used to predict cardiovascular and mortality risk[5–8]. While these models offer valuable epidemiological insights at the population level, they rely on manually extracted features (i.e., PSG annotations or diagnostic labels) and simplified categorical classification, which can overlook rich, nuanced physiological patterns in raw EEG, ECG, and respiratory data. This limitation in traditional approaches highlights the need for methods that can learn directly from raw physiological data to uncover subtle latent indicators that traditional summary metrics (e.g., sleep efficiency) may overlook.

Recent advances in deep learning have demonstrated the potential for automated feature extraction from physiological signals. Supervised deep learning models have been used to predict cardiovascular outcomes directly from physiological signals. Supervised models using ECG data have achieved high accuracy for cardiovascular risk stratification [9–11]. For example, Khurshid et al.[9] developed a deep learning method to predict incident atrial fibrillation using ECG data, while Kany et al.[11] and Al-Alusi et al.[10] used ECG to predict numerous types of cardiovascular events. However, these approaches typically rely on a single modality of input (e.g., ECG), require large, well-annotated datasets, and have limited generalizability across diverse populations. The potential of PSG in the context of cardiovascular disease (CVD) risk profiling remains largely underexplored.

Self-supervised learning has emerged as a powerful alternative for extracting meaningful representations from unlabeled physiological time series by leveraging inherent temporal and structural dependencies. In sleep research, self-supervised frameworks have shown promising results in both sleep stage classification and sleep disorder detection. Notably, recent models based on contrastive or predictive coding using only single-modal EEG signals have achieved accuracy comparable to supervised methods in sleep stage classification [12,13]. More recently, self-supervised learning has been extended to multimodal PSG data, incorporating EEG, ECG, and respiratory signals to learn joint latent spaces that enhance sleep-related classification tasks [14,15]. Beyond sleep staging, growing evidence suggests that physiological signals encode subtle features indicative of long-term cardiovascular health [16,17]. For instance, Friedman et al.[18] proposed an unsupervised ECG embedding approach that produces scalar risk scores predictive of CVD outcomes, demonstrating the broader potential of self-supervised models for risk stratification. However, despite these advances, no prior study has systematically explored the use of unsupervised multimodal PSG representations for cardiovascular risk profiling, leaving a significant gap in understanding the latent disease-relevant patterns embedded in sleep physiology.

In this study, we introduce a self-supervised, multimodal framework that embeds overnight PSG signals, including EEG, ECG, and respiratory waveforms, into a shared latent space for cardiovascular risk profiling. Using data from over 4,000 participants in the Sleep Heart Health Study (SHHS), we constructed projection vectors for cardiovascular outcomes by contrasting individuals with and without key cardiovascular or mortality outcomes. Each participant's physiological profile was then summarized using a set of interpretable projection scores, reflecting alignment with different disease phenotypes. We evaluated whether combining these scores with conventional risk factors, such as age, sex, and the Framingham Risk Score (FRS), enhances predictive performance across a broad set of cardiovascular outcomes. To evaluate generalizability, we validated our findings on an independent cohort from the Wisconsin Sleep Cohort (WSC). This work aimed to unlock the latent value of PSG signals by uncovering hidden patterns underlying the data for disease risk profiling. It offers a scalable and interpretable framework for individual-level cardiovascular risk stratification without relying on labels.

## 2 Methods

### 2.1 Datasets

We utilized the SHHS dataset [21,22], a community-based, prospective cohort study, to develop and evaluate the proposed self-supervised learning framework. SHHS provides comprehensive PSG recordings including EEG, ECG, respiratory signals, blood pressure, and extensive biographical and clinical information. The dataset is structured into SHHS1 (baseline), SHHS follow-up, and SHHS2. SHHS2 is a second in-home PSG that was performed approximately 5 years later than SHHS1 for a subset of participants (~50%), while the follow-up includes some CVD data without PSG signals between SHHS1 and SHHS2. The data on CVD outcomes from the SHHS follow-up and SHHS2 examinations make SHHS a uniquely valuable resource for studying both current and long-term health risks associated with sleep patterns. The data used in this study were obtained from the SHHS via the National Sleep Research Resource (NSRR)[1]. Access was granted through a data use agreement, and all analyses were conducted in compliance with NSRR's data use policies and institutional ethical standards.

---
[1] https://sleepdata.org/datasets/shhs

A detailed flowchart of the cohort selection and preprocessing pipeline is provided in Figure 1. Among the 6,441 enrolled participants, 5,804 had valid overnight PSG recordings. We excluded individuals with poor signal quality or missing key clinical variables required for downstream analysis, such as body mass index (BMI), total cholesterol, or high-density lipoprotein (HDL) levels. After applying these criteria, 4,398 participants were retained for analysis. The final cohort was randomly split into training/validation (80%) and held-out test (20%) subsets, with 880 participants in the held-out test set, a substantial sample for evaluating associations between sleep physiology and clinical outcomes. EEG and ECG signals were sampled at 125 Hz, while respiratory signals were sampled at 10 Hz. To streamline the modeling pipeline, no additional preprocessing was applied to the raw PSG signals. Model training and hyperparameter tuning were performed on the training and validation subsets, while final performance evaluation was conducted exclusively on the test subset.

For downstream analyses, both prevalent and incident disease outcomes were considered. Prevalent disease labels were derived from clinical assessments recorded at SHHS1. To define incident hypertension, we followed the definition of SHHS. We identified participants who were normotensive at SHHS1 but reported antihypertensive treatment or diagnosis during follow up. CVD mortality was defined as death due to cardiovascular causes within five years (1,825 days) following the SHHS1 recording. Incident AF was considered present if AF was identified on a 12-lead ECG obtained at the second SHHS exam or was adjudicated by the parent cohorts at any time between the baseline PSG and the final follow-up date for AF ascertainment of June 30, 2006. We selected myocardial infarction (MI), right bundle branch block (RBBB), congestive heart failure (CHF) and atrial fibrillation (AF) as representative prevalent cardiovascular conditions for evaluation.

For external validation, we utilized the Wisconsin Sleep Cohort (WSC) [19]. This external validation step allowed us to evaluate the generalizability and robustness of our findings across an independent cohort with distinct demographic characteristics and PSG acquisition protocols, thereby strengthening the rigor and clinical relevance of our framework. The WSC includes overnight, in-laboratory polysomnography conducted at the University of Wisconsin–Madison Clinical and Translational Research Core (CTRC), with a baseline sample of approximately 1,500 Wisconsin state employees. The details of the WSC dataset lie in Figure 2. Participants have undergone repeat assessments at roughly four-year intervals, providing a rich dataset for long-term outcome analysis.

We used five clinically relevant outcomes from the WSC dataset: incident cardiovascular disease, incident hypertension, prevalent hypertension, prevalent CVD, and prevalent coronary artery disease. For incident outcomes, we define cases as individuals who were free of the condition at their baseline sleep study but developed the disease at a subsequent follow-up visit. This enables us to evaluate the predictive value of our physiological embeddings in identifying individuals at risk before disease onset. Prevalent cases are defined based on self-reported or physician-diagnosed conditions present at the time of the sleep study.

## 2.2 Model Development and Training

### 2.2.1 Self-supervised learning framework

Motivated by recent work on self-supervised learning for sleep stage classification using PSG signals[12,14], we utilized a Residual–Transformer–based self-supervised framework to model the relationship between PSG recordings and both prevalent and incident disease outcomes.

Our model employs a self-supervised paradigm to encode 30-second segments of physiological signals into 256-dimensional latent representations, facilitating efficient feature extraction for downstream analyses. By removing the need for manual annotation during the encoding phase, this approach is particularly suitable for large-scale clinical cohorts. Model training was conducted using Python 3.9 on a single NVIDIA RTX 3090 GPU.

The backbone architecture of our model is illustrated in Figure 4. The whole backbone is comprised of residual blocks and transformer blocks to better capture short-term patterns and long-term dependencies. Each input sample $X \in R^m$ is represented as a time-series vector of length $m$, where m is the product of the duration (in seconds) and the sampling rate. The input signal is then partitioned into non-overlapping patches $Patch \in R^{n \times d}$ using a series of residual blocks, where $n$ denotes the number of patches and $d$ is the embedding dimension accepted by the encoder. We then apply position embeddings and input patches into transformer encoders to acquire an embedding $E \in R^{n \times d}$; the output dimension is the same as the input dimension. This hybrid architecture enhances training stability and enables joint modeling of short-term and long-term dependencies.

For self-supervised learning, we trained the model to infer masked segments from the remaining parts of the PSG signals, thus encouraging the learning of informative latent representations. Briefly, we applied random masking to the non-overlapping $Patches \in R^{n \times d}$. A masking function $M \in R^n$ is applied to identify the masked patches. Within $M$, a value of 1 is assigned for masked tokens and 0 for unmasked tokens, effectively delineating the regions of the time series data that are concealed and revealed.

The loss function minimizes the discrepancy between the reconstructed masked segments and the corresponding unmasked segments. The decoder $d_\phi$ has a similar structure to the encoder, with a minor difference in depth. Here, the $E^M$ denotes the embedding features of signals after random masking, and $\hat{E}$ denotes the embedding features of the complete input signals. The loss function we used is as follows:

$$L_{Similarity} = D\left(\hat{E}, d_\phi(E^M)\right) \quad (1)$$

$$D = Cosine\ Similarity \quad (2)$$

$$Z_{list} = [d_\phi(E^{M_1}), d_\phi(E^{M_2})..d_\phi(E^{M_n})] \quad (3)$$

We performed random masking multiple times per input segment. The similarity loss $L_{Similarity}$ was computed as the mean over several runs, using 20–30 random mask permutations per segment to ensure robust representation learning. The purpose of this self-supervised training was to extract the key patterns of input signals.

To mitigate representational collapse (i.e., the failure to learn meaningful features), we incorporated a Total Coding Rate (TCR) penalty as proposed by Ma et al.[20]. The TCR loss is defined as follows, where the latent representation is derived from the previously described list $Z_{list}$.

$$L_{TCR} = \frac{1}{2} logdet\left(I + \frac{d}{b\epsilon^2}ZZ^T\right) \quad (4)$$

The model was optimized in the self-supervised phase by these two loss functions.

### 2.2.2 Disease Vector Derivation

An illustration of this procedure is shown in Figure 3, where the embeddings from two groups are depicted in a 2D space. The centroids of disease-positive and disease-negative subjects are computed separately, and the disease vector $\vec{v}_{disease}$ is defined as the directional difference between them. We derived disease vectors by computing the directional difference between the centroids of disease-positive and disease-negative samples. This conceptually straightforward approach enables the construction of numerous disease-specific vectors across multiple PSG signal modalities, providing substantial flexibility in representation learning. These vectors were computed during the training phase and subsequently evaluated on the testing subset to assess their effectiveness to detect disease-related deviations in latent signal representations.

The centroids were calculated as the mean embeddings of physiological signals (e.g., EEG, ECG, respiratory signals) from individuals with a given disease and from those without, respectively. Specifically, for each cardiovascular disease and outcome, we aggregated the embeddings of subjects with positive labels and computed their average, and did the same for those with negative labels, yielding a pairwise centroid difference vector representing the disease-specific direction in the embedding space. Formally, the disease vector is computed as:

$$\vec{v}_{\text{disease}} = \mu_{\text{positive}} - \mu_{\text{negative}} \quad (5)$$

where $\mu_{\text{positive}}$ and $\mu_{\text{negative}}$ represent the average embeddings of disease-positive and disease-negative subjects, respectively.

### 2.2.3 Disease or Outcome Score Projections

During the testing phase, disease scores for each physiological modality were computed by projecting the corresponding signal embeddings onto pre-defined disease vectors. This scalar projection score provides a continuous valued score interpretable as the degree of disease-related deviation in the latent space and reflects how closely a subject's physiological signals align with disease-associated patterns.

To ensure consistency across subjects and modalities, all embedding vectors were normalized to have unit norms. Disease vectors were also normalized to unit norm prior to projection. This transformation enables scalar representations from high-dimensional embeddings, facilitating downstream modeling (e.g., logistic regression) and enhancing the interpretability of disease-associated deviations in the latent space.

$$Score_{signal} = \frac{Signal * V_{disease}}{V_p} \quad (6)$$

For each physiological signal modality, we computed subject-level projection scores by averaging the top three highest segment-level projections onto disease-specific latent vectors. This approach accounts for the possibility that only a few segments may exhibit disease-relevant patterns, particularly in patients with sparse or subtle manifestations. The resulting scores quantify the alignment between an individual's physiological representation and the latent direction associated with a specific disease phenotype.

### 2.2.4 Predictive Model Development and Statistical Analysis

We conducted group-wise statistical comparisons to assess potential baseline differences in covariates between the disease and non-disease groups. For continuous variables, such as age and body mass index (BMI), we applied the Kruskal–Wallis test, a non-parametric alternative to ANOVA suitable for non-normally distributed data. For categorical variables, including sex and disease prevalence, we used the chi-square test. These preliminary analyses helped ensure that subsequent modeling results were interpreted in the context of observed group differences. During the training phase, we fitted logistic regression models using a combination of traditional covariates (sex, BMI, and age), our projection scores in different modalities, and the Framingham Risk Score (FRS) for comparison[21,22]. We adopted the laboratory-based version of the FRS, which incorporates clinical measurements such as blood pressure and cholesterol levels. FRS score is based on gender, age, total cholesterol, HDL, systolic blood pressure, medication for hypertension, smoker and diabetes status.

We tested various combinations of these predictors to evaluate the incremental value of our projection scores and to assess their complementarity with established clinical risk factors. Odds ratios were computed with 95% confidence intervals to assess the strength and significance of associations between projection scores and disease outcomes. Each projection score was entered into a multivariable logistic regression model alongside traditional covariates, including age, sex, and BMI. This allowed us to evaluate whether the projection-based disease vectors captured complementary risk information beyond conventional demographic and clinical factors. All odds ratios reported reflect the independent effect of each variable when adjusted for the others in the model. The trained models were subsequently applied to the held-out testing set to evaluate the discriminative power of our projection-based disease vectors. Our goal was to determine whether the learned disease scores provide complementary information beyond predefined medical covariates and established clinical scores. We used the area under the receiver operating curve (AUC) as the metric to evaluate performance. We considered that $p < 0.05$ indicated statistical significance in our analysis. Model fitting and statistical analysis were performed using the Python package statsmodels v0.14.4, which provides robust estimation of *p*-values and odds ratios.

## 3 Results

### 3.1 Cohort Characteristics

Tables 1 and 2 present the demographic and clinical characteristics of participants from the SHHS1 and WSC cohorts, respectively. SHHS1 included 4,398 individuals (3,518 in training/validation and 880 in testing), while WSC included 1,093 individuals (1093 in testing). Across both cohorts, training and test sets were well matched on key demographic variables such as age, sex, body mass index (BMI), and systolic blood pressure (SBP). The prevalence of baseline conditions, including various types of cardiovascular diseases, was also similar between training/validation sets and testing sets in SHHS1.

## 3.2 Projection Scores Across modalities and cardiovascular outcomes

To assess the discriminative power of the learned embedding scores, we analyzed their distribution across several CVD conditions. We first visualized the distribution of projection scores using violin plots, stratified by disease status (positive vs. negative) for SHHS1 subjects with complete prevalent and incident CVD outcomes in follow-up cohort. Figures 5, 6, and 7 show the projection score distributions across eight CVD outcomes (diseases and mortality) for ECG, EEG, and respiratory signals respectively. Among the three modalities, ECG-based scores (Figure 5) exhibited the most pronounced group differences where individuals in the negative outcome groups consistently demonstrated elevated projection scores compared to controls ($p < 0.001$ across most phenotypes). EEG-based scores (Figure 6) showed moderate differences in cardiovascular outcomes, albeit with subtler margins than ECG (with several comparisons reaching statistical significance). Respiratory signals (Figure 7) also revealed distinguishable differences across different prevalent and incident diseases, though with greater overlap between groups.

## 3.3 Predictive Performance in Outcome Classification

To evaluate the predictive utility of sleep-derived projection scores, we trained logistic regression models using different combinations of projection scores, demographic variables, and the Framingham Risk Score (FRS). Performance was assessed on held-out test sets within both the SHHS1 and WSC cohorts. Summary results are presented in Table 3 and 4.

Across both datasets, projection scores derived from overnight PSG signals demonstrated strong discriminative capacity for cardiovascular outcomes. In the SHHS test set, models that combined projection scores with the FRS consistently yielded the highest AUCs, achieving 0.965 for atrial fibrillation, 0.861 for congestive heart failure, 0.762 for myocardial infarction, and 0.854 for cardiovascular mortality. For hypertension, the combined model reached an AUC of 0.780 for prevalent cases and 0.607 for incident cases, outperforming both clinical baselines and projection-only models (based on EEG, ECG, or respiratory signals used individually or in combination).

Projection-only models, trained without PSG diagnostic labels, also captured disease-relevant features, hidden patterns correlated with CVD inside PSG signals. For example, ECG-based projection scores alone yielded near-perfect discrimination of right bundle branch block (AUC = 0.997) and atrial fibrillation (AUC = 0.961). Also, EEG-derived scores achieved AUCs above 0.60 for hypertension and mortality.

In external validation using the WSC cohort, the additive value of the projection scores was preserved. Models that integrated both projection scores and the FRS achieved high AUCs for coronary artery disease (AUC = 0.807), incident cardiovascular disease (AUC = 0.753) and hypertension (AUC = 0.805), outperforming models based solely on FRS or demographic features.

We further observed modality-specific trends. ECG-based models performed best in outcomes directly related to cardiac structure and rhythm (AF, CHF), while EEG scores contributed more to vascular conditions such as hypertension and mortality. Respiratory features alone were less predictive, but enhanced performance in composite models, such as in CHF (ECG–Resp, AUC = 0.787) and MI (ECG–Resp, AUC= 0.676).

### 3.4 Interpretability Assessment

To further evaluate the interpretability of the learned projection scores, we computed adjusted odds ratios (aORs) using logistic regression models adjusted for traditional demographic risk factors, including age, sex, and BMI. Figure 8 represents a forest plot showing the associations between projection scores derived from individual physiological modalities (EEG, ECG, and respiratory signals) and disease outcomes. The majority of modality–disease associations were statistically significant ($p < 0.005$), as indicated by nonoverlapping confidence intervals. Notably, ECG-derived projection scores exhibited strong associations across all outcomes, with aORs exceeding 1.5. While EEG-based projection scores showed slightly weaker associations with cardiovascular outcomes, they still demonstrated meaningful predictive value, for example, an aOR of 1.57 for CVD mortality and 1.51 for prevalent hypertension.

Projection scores derived solely from respiratory signals were also significantly associated with several outcomes, suggesting that physiological risk information is embedded within respiratory dynamics as well.

### 3.5 Illustration of the Patient Report Card

To illustrate clinical interpretability, we present a patient-specific cardiovascular risk card as an illustrative example. The average patient score is calculated based on the SHHS dataset, while the data from one patient in SHHS with hypertension. In this illustrative example (Figure 9), an ECG from a patient is projected onto the phenotype vectors for selected cardiovascular diseases and outcomes, and the positions relative to the whole cohort along vectors from the disease-negative to disease-positive centroids are reported. This workflow can also be applied to EEG and respiratory signals, providing an easy-to-use tool for clinicians.

**Discussion**

We introduced a self-supervised, multimodal framework for deriving interpretable physiological embeddings from overnight PSG recordings. Unlike traditional approaches that rely on labor-intensive PSG annotation and simplified categorization, our method projects raw EEG,

ECG, and respiratory data into a shared latent space, capturing modality-specific and complementary features relevant to cardiovascular health, beyond established risk factors such as the FRS. In both the training and external validation cohorts, the projection scores demonstrated strong predictive values for cardiovascular outcomes. For example, combined with the FRS score, the projection scores can achieve an AUC of more than 0.75 in most cardiovascular outcomes. Notably, this performance remained consistent in the external dataset, underscoring the generalizability and clinical utility of the embeddings.

Our framework advances beyond traditional sleep architecture analysis by learning physiological representations from PSG predictive of downstream clinical outcomes. While prior self-supervised approaches have focused primarily on sleep stage classification[12,15], our approach generates projection scores that link learned features to a broader range of cardiovascular outcomes. Notably, when integrated with the FRS, our projection scores achieved an AUC of 0.854 for CVD mortality and exceeded 0.75 across most cardiovascular conditions. Even in more subtle scenarios, such as incident hypertension, the combined model maintains meaningful discriminative power (AUC = 0.607), outperforming the FRS alone. These findings suggest that projection scores capture risk signals not fully reflected in traditional models, offering added value for early and individualized risk stratification.

We found that each modality contributed uniquely to predictive performance. ECG-derived scores demonstrated the most robust and consistent performance, with AUCs exceeding 0.96 for atrial fibrillation and RBBB, and odds ratios greater than 2.0 for CVD mortality, MI, and CHF. EEG-derived scores were predictive of increased risk for CVD mortality and incident hypertension. While respiratory-derived scores were less discriminative on their own, they added incremental value in multimodal models, improving CHF prediction when combined with ECG. These associations remained stable in an external validation cohort, highlighting the generalizability of the learned representations.

A particularly notable finding was the consistent association between EEG-derived projection scores and multiple cardiovascular outcomes, suggesting their potential as digital biomarkers for cardiovascular risk. Previous studies have linked EEG features to specific cardiovascular conditions such as stroke. For example, Torres et al. [23] found that EEG changes occur whenever the patient has CVDs. Furthermore, Lamat et al.[24] and Niu et al.[25] used EEG features (e.g., mean value in time-domain and power spectral density in frequency-domain) to predict different types of strokes. More recently, Zhou et al.[26] proposed a sleep depth index derived from EEG signals, demonstrating associations with hypertension and CVD mortality. However, their approach relies on traditional sleep stage metrics derived from EEG. Our approach extends this by using self-supervised learning to uncover broader associations between EEG signals and cardiovascular risk. EEG-derived projection scores were significantly associated with hypertension, CVD mortality, and incident hypertension, highlighting their relevance across a wide spectrum of outcomes.

This study has several limitations. First, although we demonstrated generalizability on the WSC cohort, our training data (SHHS1) primarily consisted of older, white adults from the United States, which may limit the model's applicability to more diverse populations. Second, while our current approach enables segment-level attribution and temporal localization of outcome-relevant patterns, the resolution of this localization remains relatively coarse.

Although this study has several limitations, our findings have several implications for clinical practice. This framework has the potential to transform the clinical utility of PSG studies. By extracting meaningful physiological embeddings from routine overnight recordings without requiring manual annotations, it enables early and individualized risk stratification for a wide range of cardiovascular outcomes. The integration of projection scores with established risk models like the FRS consistently improved predictive performance, suggesting that these embeddings capture complementary physiological signals not reflected in traditional clinical variables. This could support more personalized decision-making in preventive cardiology, helping identify high-risk individuals who may benefit from early intervention. Moreover, the framework's interpretability and scalability make it highly amenable to integration into clinical workflows, thereby extending the utility of sleep studies beyond traditional sleep disorder diagnostics to comprehensive systemic health surveillance. By simply uploading patients' demographic data and physiological signal recordings, the system can automatically compute projection scores alongside FRS, facilitating early identification of high-risk individuals and enabling proactive preventive care. Furthermore, model inference is computationally efficient, requiring only a single GPU with approximately 10GB of VRAM, and the end-to-end processing time for each patient remains under 5 seconds, ensuring practical deployment in time-sensitive clinical settings.

In summary, we present a scalable, interpretable, self-supervised framework that learns physiological embeddings from overnight PSG recordings, enabling individualized and multimodal risk stratification for a broad range of cardiovascular outcomes. The integration of the projection scores with established risk factors consistently improves predictive performance, supporting the added clinical value of our approach. The framework's interpretability and scalability make it well-suited for integration into routine clinical workflows, offering the potential for earlier identification of individuals at elevated risk and more personalized strategies for disease prevention and management.

**Acknowledgment**: This research was supported by grants from the National Science Foundation (2052528). The Sleep Heart Health Study was supported by National Heart, Lung, and Blood

**Disclosure Statement**

Financial disclosure: The authors declare that they have no financial relationship or financial conflicts of interest to disclose.

Non-financial disclosure: The authors declare that they have no non-financial conflicts of interest to disclose.

**Figure Legends**

Figure 1: Flowchart of Polysomnographic (PSG) Recordings Selection from the Sleep Heart Health Study (SHHS1).

Figure 2: Flowchart of Polysomnographic (PSG) Recordings Selection from the Wisconsin Sleep Cohort (WSC). n = number of recordings, SD = standard deviation

Figure 3: Visualization of disease vector construction in the embedding space. Each dot represents the embedding of physiological signals from an individual subject, with disease-positive samples shown in orange and disease-negative samples in blue. The centroids of the two groups, $\mu_{positive}$ and $\mu_{negative}$, are computed, and the disease vector $\vec{v}_{disease}$ is defined as their difference. This vector captures the direction of disease-related variation in the latent space.

Figure 4: Overview of the proposed backbone architecture, comprising residual and transformer blocks for capturing both local and global signal dependencies.

Figure 5: Violin plots showing projection score distributions from the ECG modality across control and disease groups for eight clinical outcomes. Wider sections represent higher density of patients at a given score level.

Figure 6: Violin plots of disease-specific projection scores derived from EEG embeddings, stratified by disease status.

Figure 7: Violin plots illustrating projection score distributions from respiratory signal embeddings across disease and control groups.

Figure 8: Forest plot of odds ratios (ORs) and 95% confidence intervals (CIs) derived from logistic regression models using projection scores from EEG, ECG, and respiratory modalities for each disease outcome. All models are adjusted for age, sex, and BMI. Significant associations ($p < 0.005$) are marked with triple asterisks (***).

Figure 9: Example of a patient-specific cardiovascular risk card based on projection scores. The patient's ECG-derived scores are compared with population-level averages from the SHHS cohort across multiple disease outcomes, enabling individualized physiological interpretation.

Table 1: Demographic and clinical characteristics of SHHS1

| Characteristic | Training set (N=3518) | Test set (N=880) |
| --- | --- | --- |
| Age, mean (SD) | 63.65 (10.79) | 63.35 (11.14) |
| Male, n (%) | 1664 (47.32%) | 433 (49.20%) |
| BMI, mean (SD) | 27.98 (4.89) | 28.19 (4.90) |
| Systolic Blood Pressure, mean (SD) | 126.88 (19.41) | 125.49 (18.48) |
| Hypertension, n (%) | 2381 (67.7%) | 589 (66.9%) |
| Congestive Heart Failure, n(%) | 83 (2.4%) | 13 (1.5%) |
| Atrial Fibrillation, n (%) | 26 (0.7%) | 13 (1.5%) |
| Myocardial Infarction, n (%) | 196 (5.6%) | 49 (5.6%) |
| Right Branch Block, n (%) | 63 (1.8%) | 21 (2.4%) |
| Cardiovascular Mortality, n (%) | 78 (2.2%) | 11 (1.3%) |
| Incident Hypertension, n (%) | 143 (4.1%) | 41 (4.7%) |
| Incident Atrial Fibrillation, n (%) | 216 (6.1%) | 60 (6.82%) |

SD: standard deviation; n: number of participants.

Table 2: Demographic and clinical characteristics of WSC.

| Characteristic | Test Set (N=1093) |
|---|---|
| Age, mean (SD) | 56.40(8.13) |
| Male, n (%) | 572 (52.33%) |
| BMI, mean (SD) | 31.68 (7.18) |
| Systolic Blood Pressure, mean (SD) | 127.17 (15.49) |
| Hypertension, n (%) | 361 (33.03%) |
| Coronary Artery Disease, n (%) | 69 (6.31%) |
| Cardiovascular Disease, n (%) | 105 (9.61%) |
| Incident Hypertension, n (%) | 86 (7.87%) |
| Incident Cardiovascular Disease, n (%) | 26 (2.38%) |

SD: standard deviation; n: number of participants.

Table 3: Comparison of Area under the Curves (AUCs) predicting cardiovascular outcomes using different modalities and composite scores (Data from SHHS1)

| Modality | RBBB | CHF | AF | HTN | INAF | INHP | MI | Mortality |
|---|---|---|---|---|---|---|---|---|
| EEG | 0.526 | 0.517 | 0.470 | 0.635 | 0.615 | 0.583 | 0.504 | 0.615 |
| ECG | **0.997** | 0.737 | 0.961 | 0.616 | 0.711 | 0.552 | 0.654 | 0.711 |
| Resp | 0.438 | 0.624 | 0.602 | 0.580 | 0.615 | 0.476 | 0.641 | 0.615 |
| EEG-ECG | **0.997** | 0.696 | 0.932 | 0.671 | 0.735 | 0.570 | 0.661 | 0.735 |
| EEG-Resp | 0.473 | 0.582 | 0.546 | 0.647 | 0.635 | 0.564 | 0.612 | 0.635 |
| ECG-Resp | **0.997** | 0.787 | 0.965 | 0.634 | 0.704 | 0.538 | 0.676 | 0.704 |
| EEG-ECG-Resp | **0.997** | 0.730 | 0.932 | 0.679 | 0.720 | 0.560 | 0.681 | 0.720 |
| Baseline[1] | 0.728 | 0.800 | 0.765 | 0.654 | 0.841 | 0.564 | 0.656 | 0.841 |
| FRS Score[2] | 0.700 | 0.834 | 0.691 | 0.773 | 0.720 | 0.576 | 0.746 | 0.720 |
| FRS Score Composit[3] | 0.762 | **0.861** | **0.965** | **0.780** | 0.783 | **0.607** | **0.762** | **0.854** |
| Composit[4] | **0.997** | 0.848 | **0.965** | 0.700 | **0.854** | 0.597 | 0.711 | 0.783 |

Bolded values indicate the highest AUC for each column.

RBBB: right bundle branch block; CHF: congestive heart failure; AF: atrial fibrillation; HTN: hypertension; INHP: incident hypertension; MI: myocardial infarction; INAF: incident atrial fibrillation; Resp: respiratory signals

[1] "Baseline" includes age, sex, and BMI. [2] "FRS Score" refers to the Framingham Risk Score.

[3] "FRS Score Composite" combines EEG-ECG-Resp projection scores with the Framingham Risk Score.

[4]"Composite" models combine EEG-ECG-Resp projection scores with age, sex, and BMI.

Table 4: Comparison of Area under the Receiver-operating Characteristic Curves (AUCs) from WSC across all evaluated disease outcomes.

| Modality | Hypertension | CVD | Incident CVD | INHP | CAD |
| --- | --- | --- | --- | --- | --- |
| ECG | 0.561 | 0.560 | 0.540 | **0.548** | 0.610 |
| EEG | 0.545 | 0.579 | 0.672 | 0.490 | 0.469 |
| Resp | 0.528 | 0.529 | 0.518 | 0.532 | 0.523 |
| FRS Score[1] | 0.798 | 0.698 | 0.648 | 0.457 | 0.779 |
| Baseline[2] | 0.639 | 0.736 | 0.462 | 0.510 | 0.800 |
| FRS Score Composite[3] | **0.805** | 0.710 | **0.753** | 0.512 | 0.807 |
| Composite[4] | 0.666 | **0.754** | 0.520 | 0.480 | **0.812** |

Bolded values indicate the highest AUC for each column.

CVD: cardiovascular disease INHP: incident hypertension CAD: coronary artery disease.

[1]"Baseline" includes age, sex, and BMI. [2]"FRS Score" refers to the Framingham Risk Score.

[3]"FRS Score Composite" combines EEG-ECG-Resp projection scores with the Framingham Risk Score.

[4]"Composite" models combine EEG-ECG-Resp projection scores with age, sex, and BMI.

Figure 1

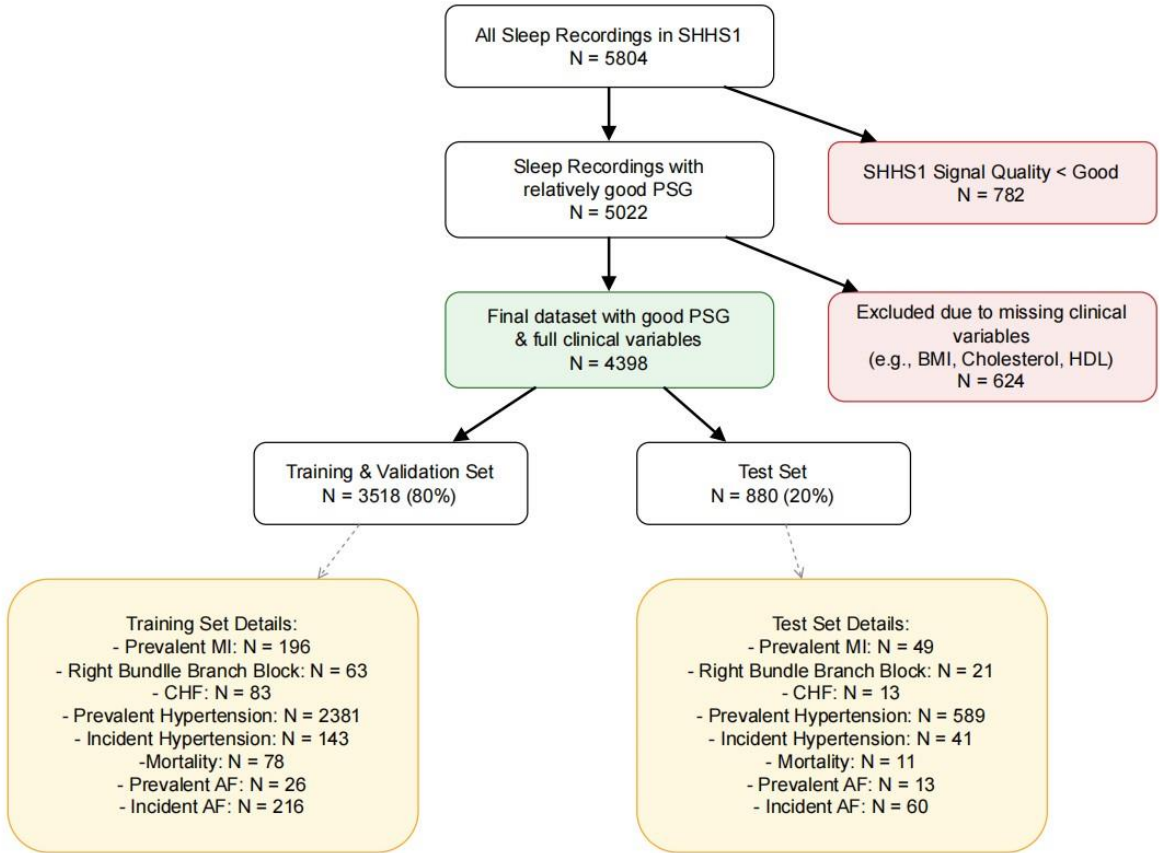

Figure 2

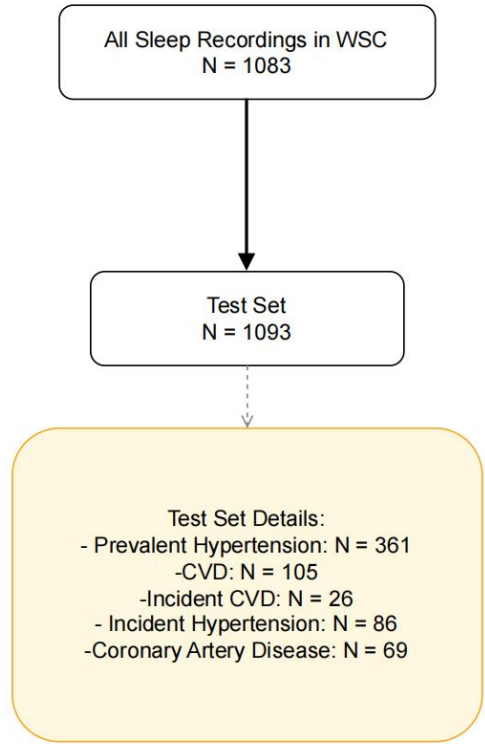

Figure 3

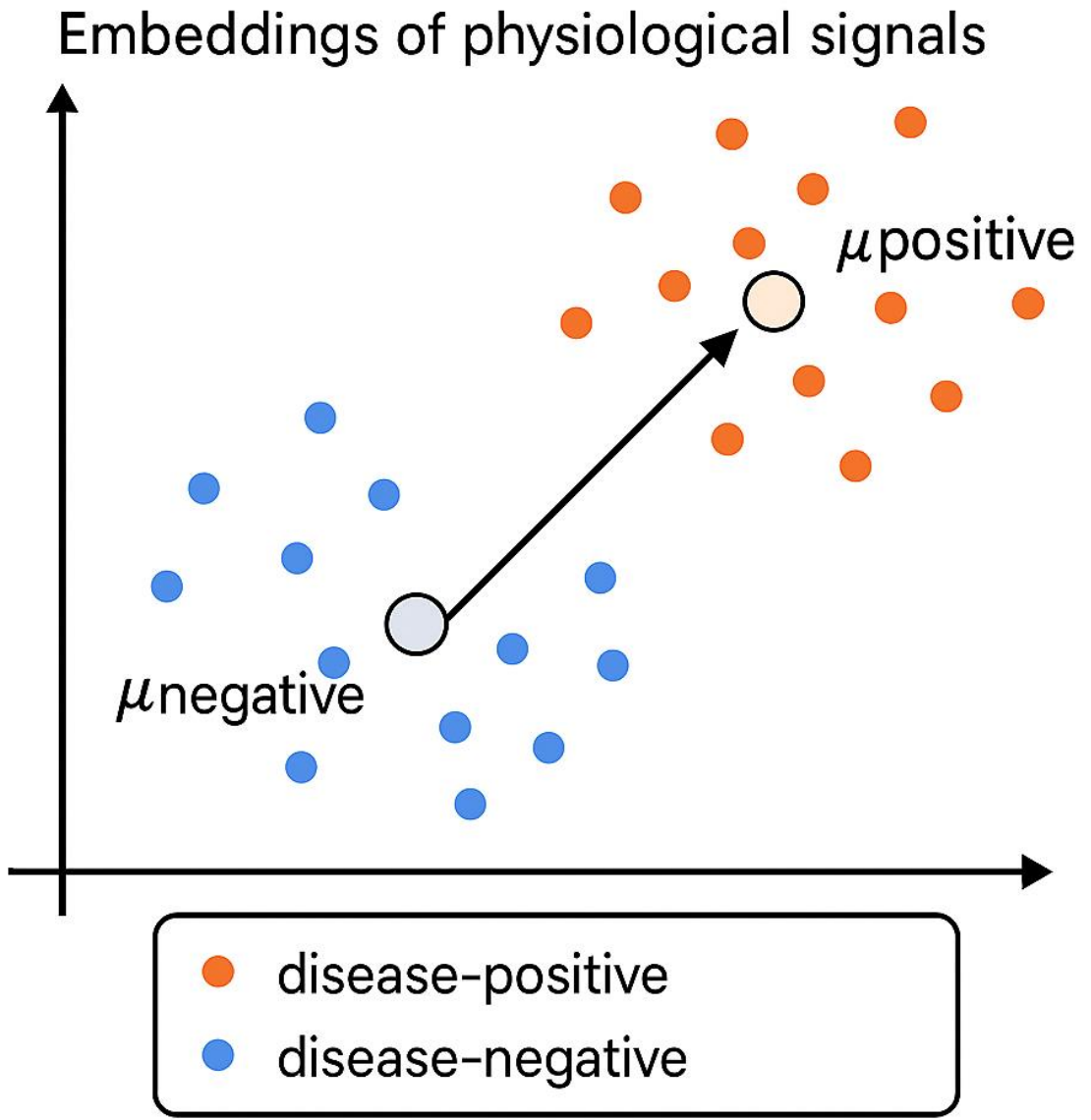

Figure 4

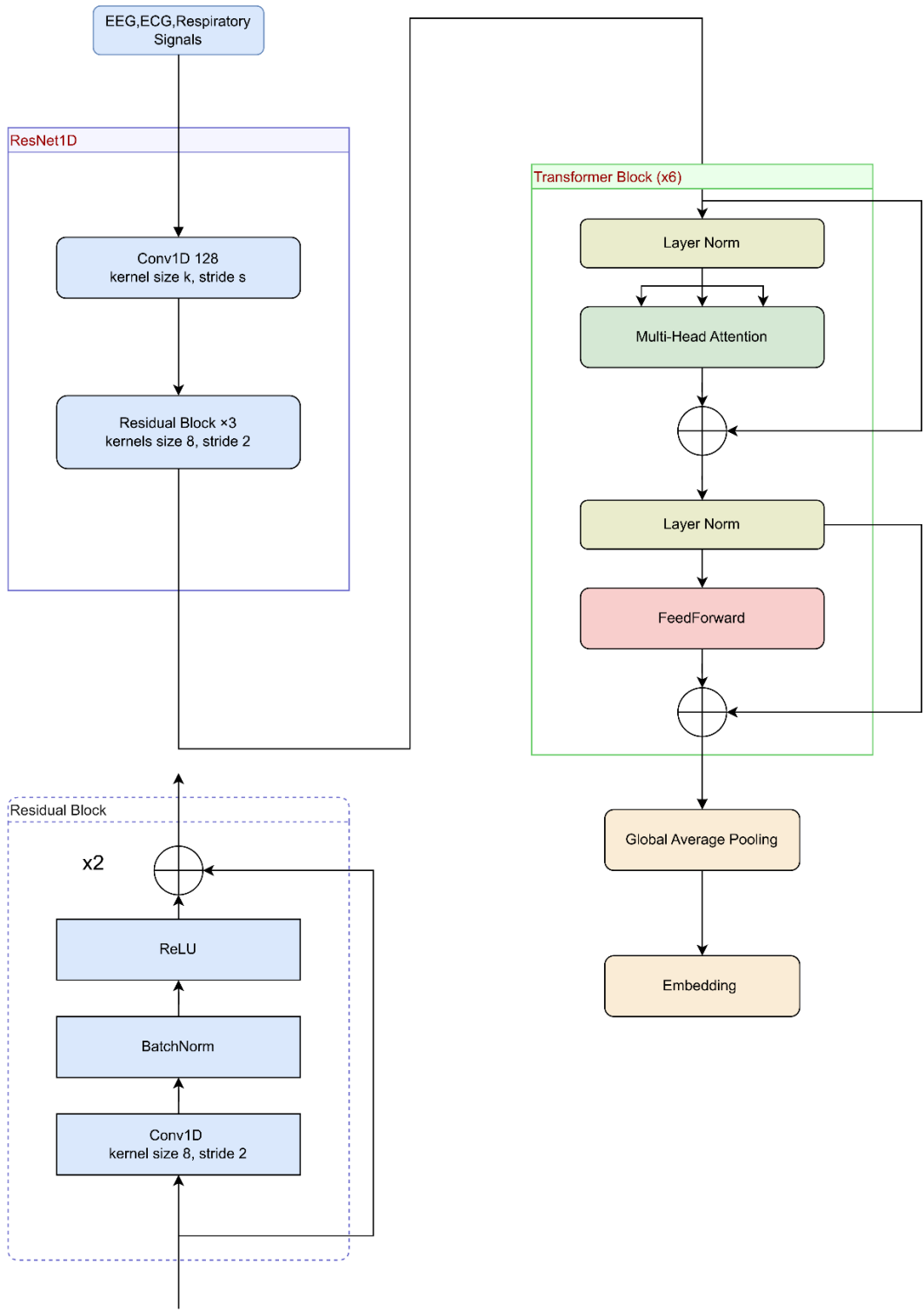

Figure 5

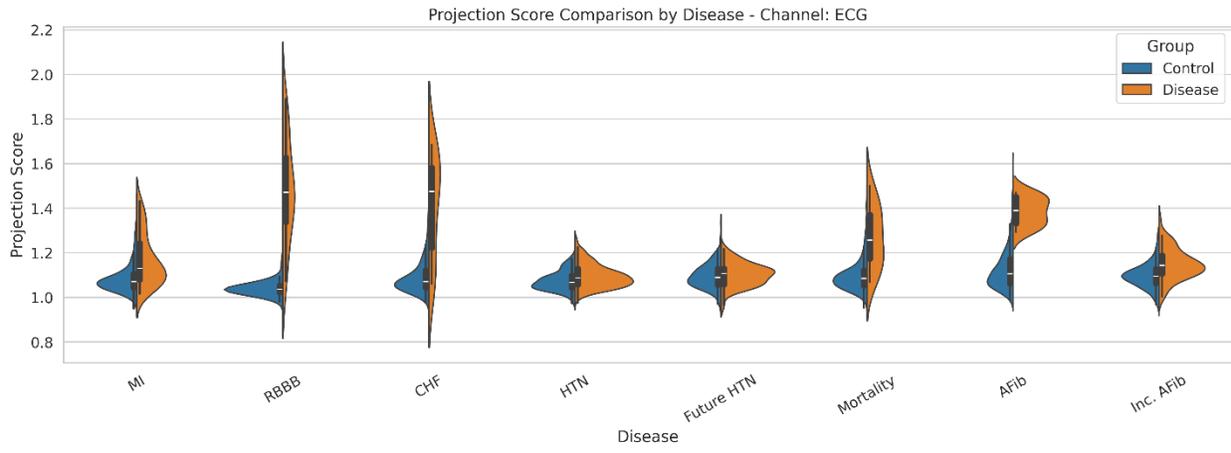

Figure 6

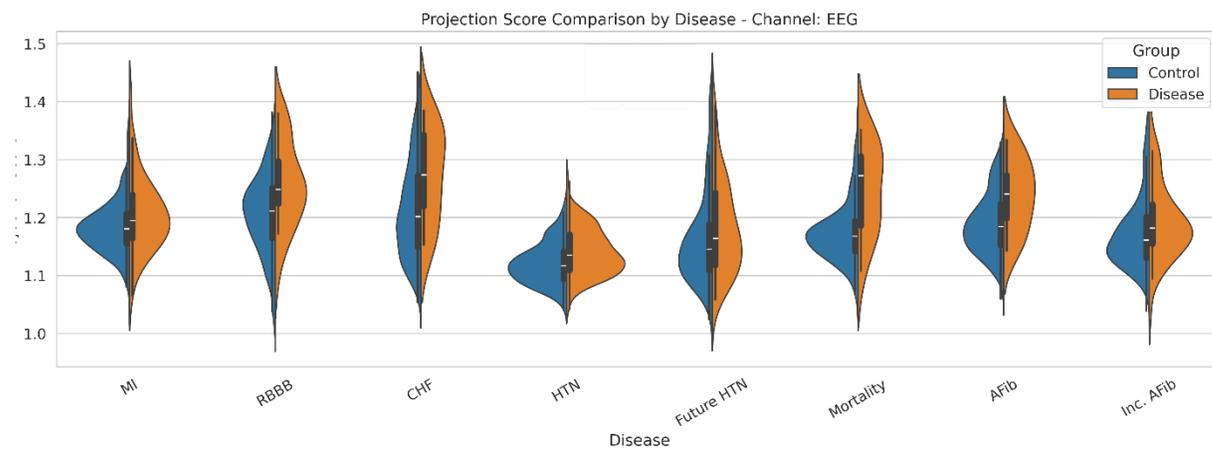

Figure 7

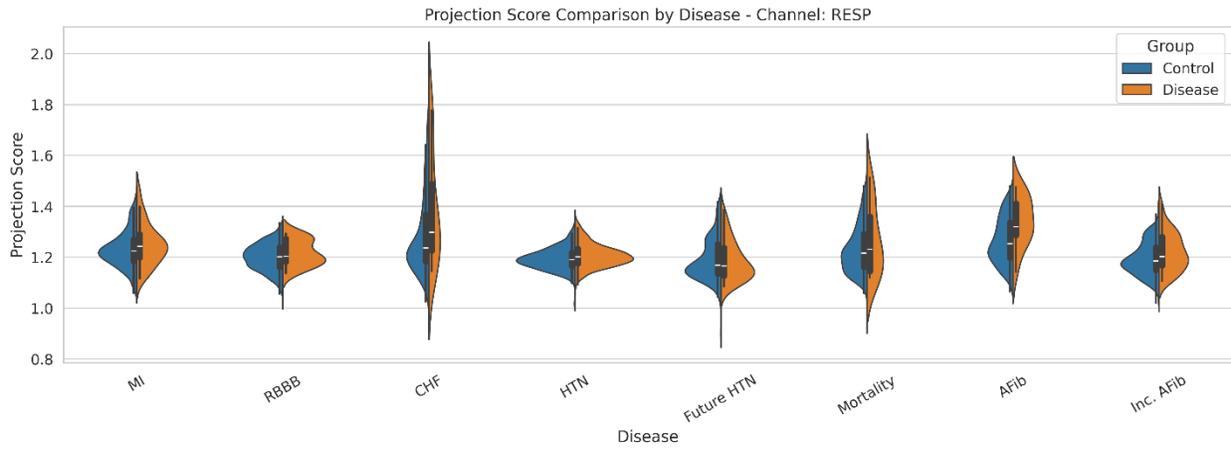

Figure 8

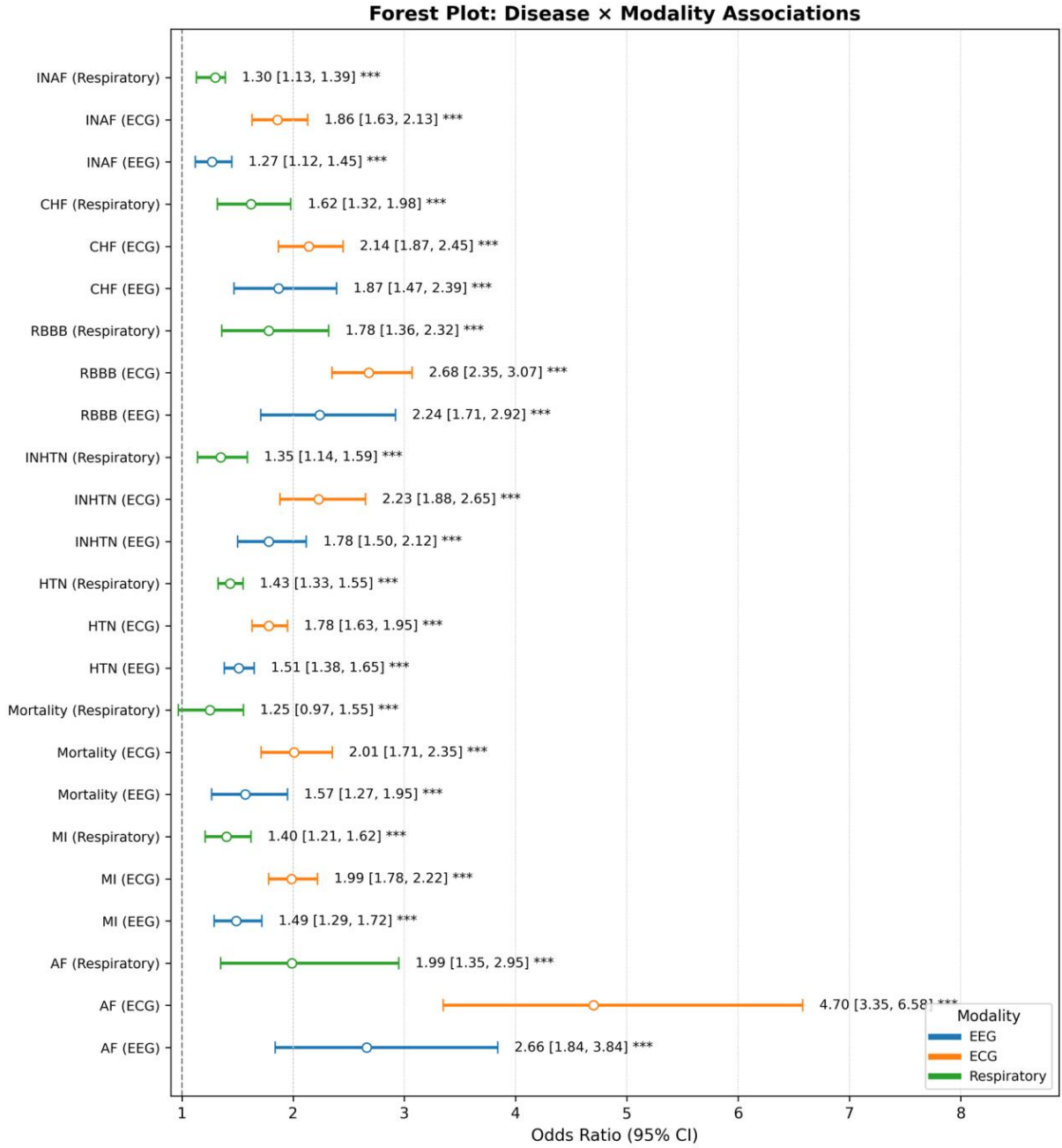

Figure 9

| Disease | Current Score | Average Score of Patients (CVD+) | Risk Status |
|---|---|---|---|
| Hypertension | 1.42 | 1.18 | Above average |
| Right Bundle Branch Block | 1.61 | 1.48 | Above average |
| Congestive Heart Failure | 1.64 | 1.49 | Above average |
| Myocardial Infarction | 1.43 | 1.16 | Above average |
| Atrial Fibrillation | 1.51 | 1.39 | Above average |
| Future Hypertension | 1.42 | 1.16 | Above average |
| Mortality within 5 years | 1.31 | 1.24 | Above average |